# Federated Learning for Energy Constrained IoT devices: A systematic mapping study

Rachid EL Mokadem, Yann Ben Maissa and Zineb El Akkaoui

{elmokadem.rachid, benmaissa, elakkaoui}@inpt.ac.ma

Telecommunications Systems, Networks and Services Lab, National Institute of Posts and Telecommunications, Rabat, 10587, Morocco.

**Abstract**

Federated Machine Learning (Fed ML) is a new distributed machine learning technique applied to collaboratively train a global model using clients' local data without transmitting it. Nodes only send parameter updates (e.g., weight updates in the case of neural networks), which are fused together by the server to build the global model. By not divulging node data, Fed ML guarantees its confidentiality, a crucial aspect of network security, which enables it to be used in the context of data-sensitive Internet of Things (IoT) and mobile applications, such as smart geo-location and the smart grid. However, most IoT devices are particularly energy constrained, which raises the need to optimize the Fed ML process for efficient training tasks and optimized power consumption. In this paper, we conduct, to the best of our knowledge, the first Systematic Mapping Study (SMS) on FedML optimization techniques for energy-constrained IoT devices. From a total of more than 800 papers, we select 67 that satisfy our criteria and give a structured overview of the field using a set of carefully chosen research questions. Finally, we attempt to provide an analysis of the energy-constrained Fed ML state of the art and try to outline some potential recommendations for the research community.

# 1 Introduction

**Context**. Machine learning (ML) has become an important and increasingly used paradigm in different applications. In the last decade, the IoT computer systems and their potential applications (e.g., smart cities, smart grids) have grown considerably, which would make them benefit from the capabilities of ML in such a large and complex context. Furthermore, widespread IoT adoption in industry and academia (e.g., via rapid prototyping platforms such as the Raspberry PI™ and Arduino™) raises expectations for data privacy preservation and efficient resource utilization in a wide range of critical applications. Therefore, in light of ML limitations for distributed systems and sensitive data, Federated Machine Learning (Fed ML) was proposed by McMahan et al. in 2016 [46] to address these constraints. The approach delegated model training tasks to client devices, which collaboratively built a global shared model that consolidated their respective local data learning while avoiding any private data from leaving its original device [32]. Since the seminal paper, Fed ML has become one of the "hot topics" in ML.

**Problem**. IoT and mobile devices have a major constraint related to energy sources, and as a result, the power consumption on these devices must be optimized for any assigned task. In particular, a machine learning algorithm is known to be a highly power-consuming multi-task process [17]. In a distributed ML setup, nodes must continually exchange data with a master node, which may drive up overhead costs for the system. FedML attempts to solve this issue by limiting the exchanged data to the local model's weights [46], trained by nodes, instead of voluminous raw data exchange. At the same time, FedML still requires improvement to enable resolving further critical challenges related to IoT and mobile device characteristics, namely, the limited resources and energy constraints [54]. As a consequence, several Fed ML works addressing these aspects have increasingly been proposed by the scientific community in the last few years. In light of this evolving literature, there is a substantial need for a comprehensive study in order to provide a clear overview of energy optimization approaches and propose new research directions for the research community.

**Contribution**. Several works have actually tackled the limitations of the original Fed ML proposal, entitled FedAvg, and proposed many optimization approaches, essentially regarding the communication load, data exchange, and other aspects, which can help to address directly or indirectly the limited energy constraint. The purpose of this paper is to conduct, to the best of our knowledge, the first Systematic Mapping Study (SMS) on Fed ML optimization for energy-constrained devices. This SMS is tasked with i.) counting and categorizing relevant primary studies published in this topic based on five research questions, ii.) analyzing and discussing the results to provide a clear understanding of recent improvements for the research community, and iii.) assisting engineers in developing innovative Fed ML solutions for IoT and mobile devices.

**Contents**. The remainder of this paper is structured as follows: First, we present related works and surveys in Section 2. Then we provide some theoretical foundations through the original FedAvg algorithm as well as the formulation of the energy optimization problem in Section 3. In Section 4, we present the method used to conduct this study, including the paper selection and filtering process, as well as the research questions (RQs).

**Table 1** Related works

| Year | Title | Type | Focus |
|---|---|---|---|
| 2020 | A Systematic Literature Review on Federated Machine Learning: From A Software Engineering Perspective. [41] | SLR | Software engineering aspects |
| 2020 | A Systematic Literature Review on Federated Learning: From A Model Quality Perspective. [40] | SLR | Model quality |
| 2020 | Federated Learning in Mobile Edge Networks: A Comprehensive Survey.[38] | Survey | General |
| 2020 | Federated Learning: A Survey on Enabling Technologies, Protocols, and Applications. [1] | Survey | Applications |
| 2020 | A Review of Privacy-preserving Federated Learning for the Internet-of-Things. [6] | Survey | Privacy |

In Section 5, we answer them and analyze the results obtained from the studied papers. We follow up with a discussion and some recommendations for research directions in Section 6. Section 7 exhibits some threats to the validity of our study. Finally, in Section 8, we conclude and outline some possible future works.

## 2 Related works

In this section, we present three surveys and two literature reviews that have been identified as being related to this work (see Table 1).[41] presented a systematic literature review on Federated Learning, from a software engineering perspective, where they covered the Federated Learning system in general, with a focus on the software development aspects and general challenges for real applications. [40], on the other hand, conducted a systematic literature review on Federated Learning from a model quality perspective, where they studied the methods for improving the quality of the Fed ML model and data. Additionally, the authors compared the model between federated and non-federated learning on the same data. Furthermore, [38] presented a survey on Federated Learning for mobile edge networks, in which they investigated the characteristics and limitations of good performance, resource allocations, communication costs, and data privacy concerns.Moreover, [1] presented a FedML survey on enabling technologies, protocols, and applications. They provided the most relevant protocols, platforms, and real-life use-cases of Federated Learning to enable data scientists to build better privacy-preserving solutions for industries; they also explored the challenges and advantages of Fed ML for real-life applications. Finally, [5] presented a survey on federated learning from a privacy preservation angle.

Although these surveys and SLRs are excellent, we think that our study tackles some aspects that were not directly addressed by them. They do not focus on the energy factor in the optimization of federated learning, except for [38], where it is not thoroughly tackled. We attempt to shed light on power consumption aspects in FedML for the IoT. As reported by Cisco in [9], IoT connections will represent more than half (14.6 billion) of all global connected devices and connections (28.5 billion) by 2022, showing their increasing pervasiveness in human lives [14]. Our personal use of smart phones and watches, which need frequent and sometimes bothersome recharging, is also a practical witness to this concern. Finally, there is also the particular case of wireless sensor networks that can be deployed in hostile environments with no possibility at all of energy replenishment.

# 3 Background

In this section, we talk about the global Federated Learning process, the FedAvg algorithm, the energy consumption problem, and some other background information.

## 3.1 Federated Learning

Federated Machine Learning is the process of developing accurate models on large-scale distributed systems made up of small devices by combining their computation power and local data [46]. The goal is to solve a class of problems that cannot be solved by a single central computer, such as those involving users' personal data, real-time computing, and on-device artificial intelligence [32].

FedML is based on a distributed architecture that involves several nodes performing training tasks on their local data and exchanging their model's parameters with a central server. The server then builds, from local models, a global aggregated model, which is equivalent to a trained model on all nodes' consolidated data. In the case of FedAvg, the global model $W_g$ is built as a weighted average of the local models $W_i$ (see equation 1).

$$W_g = \sum_i \frac{n_i}{n} W_i \quad (1)$$

The optimization of the global objective function $f$ can be expressed as the optimization of the average of local objective functions $f_i$ for all participating nodes $i = 1,…,n_i$, as given by the equation 2 [54].

$$\min_w f(w) = \min_w \frac{1}{n} \sum_i f_i(w) \quad (2)$$

where:

$$f_i(w) \coloneqq \frac{1}{k} \sum_{\xi \in D_i} l(w, \xi) \quad (3)$$

$f_i$ is defined as an average of the local loss function $l$, for each node $i$, on its local sample points, $D_i = \xi_{i1},…,\xi_{im}$ for $i \in [n]$, where $D_i$ is the local data set of the node $i$, composed of $m$ data points; $\xi_i$ and $w$ are the model parameters.

$$min_w f(w) \coloneqq min_w \frac{1}{nk} \sum_{\xi \in D} l(w, \xi) \quad (4)$$

Finally, to solve equation 4, a gradient descent method is used by each node to minimize the loss $l_i$ over its local training data $D_i$, and eventually the aggregated model $W_g$ will minimize the global objective function.

### 3.1.1 Federated Learning pseudo-algorithm

Algorithm 1 shows the idea behind *Federated Averaging* (FedAvg), proposed by [46] for Fed ML.

```
/* Run on server*/
initialize w_0;
    for each round t = 1,2, ... do
            m ←
        S_t ← (randomsetofmclients);
        for eachclient  k ∈ S_t inparallel  do
            w_{t+1} ← ClientUpdate  (k,w_t);

        w_{t+1} ← P_{k=1}^{K}  (n_k/n) w_{t+1}^{k} ;
    end
        max(C.K,1);

    end

    /* Run on client k*/
    Function ClientUpdate(k,w): B ← (split P_k
        into batches of size B); for each local
        epoch i from 1 to E do
            for batch b ∈ B do
                w ← w − μ∇l(w,b);
            end
        end
        return w to server;
    end
```

**Algorithm 1:** FedAvg pseudo-algorithm

The notations employed in the algorithm are explained underneath.

- **C** Fraction of selected clients in each round
- **K** Total number of clients
- **m** Number of randomly selected clients for each round
- $S_t$ Set of clients for each round
- $w_t$ Global model parameters at round t
- $w_t^k$ Received model parameters from client k at round t
- $n_k$ Number of data points of client k
- $n$ Total number of data points of all clients
- $P_k$ Local data-set of client k
- **B** Local data-set mini batch size to use for client training
- $B$ Set of data-set mini batches for local training
- **E** Number of training passes performed by each client before sending the update to the server.
- $\mu$ Learning rate
- $l$ Loss function
- $w$ Local model parameters

As shown in the aforementioned algorithm, the server initiates the model's parameters $w_0$, then, for each round, it determines the number **m** of participant clients to choose for training as a fraction **C** of **K** total clients. The subset of devices $S_t$ is determined randomly, and then each client device **k** receives the model's parameters $w_t$ from the server to perform the training on its respective local data set $P_k$. This training process performs a split of the local data into small batches of size **B**, and a number of **E** local epoch runs to train the local model. Finally, all selected clients compute an update of the parameters $w$, then send it back to the server, which averages them to get the new global model parameters $w_{t+1}$. This round is repeated as many times as determined by the server to reach the target performance.

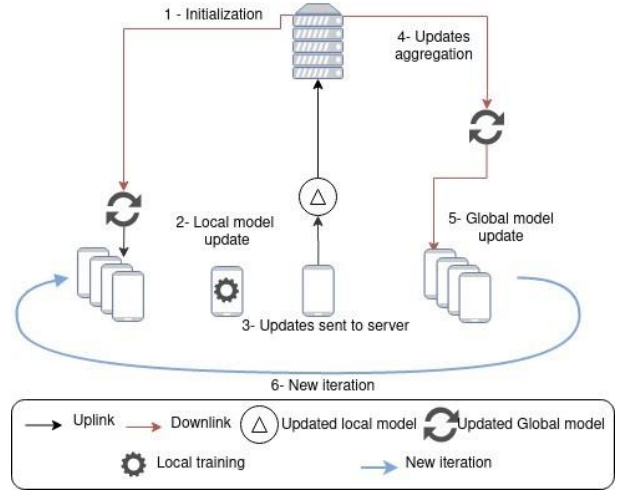

**Fig. 1** Federated Learning global schema

### 3.1.2 Federated Learning process

Fed ML architecture is composed of the client nodes and the central server. The server receives the computed updates from client devices and performs an aggregation operation to build the global model. It is then improved continuously, by running additional iterations on the nodes, to train their local models, until obtaining the desirable results.

Figure 1 globally shows the components involved in the Fed ML architecture, as well as the stages of the FedAvg algorithm execution. In each round of the training, the following operations are performed:

1. **Definition of model's structure, random initiation of parameters and selection of participating devices**: the central server must define the parameters **E**, **C** and **B** prior to start of the training, and it must select a subset of clients to participate in each round.
2. **Model Update on local data**: each selected client computes an update of the global model, by running local training iterations as many times as defined by the central server.
3. **Transmission of Local Model updates to server**: each participating device sends the computed update of the model.
4. **Aggregation of all received model updates**: the server aggregates the received updates in such a way that builds a global model.
5. **Sharing the updated global model with the devices**.

### 3.1.3 Heterogeneity

Very often, in real applications, the participant nodes in the FL have uneven resources and training data, we refer to this by *system heterogeneity* and *statistical heterogeneity* respectively [35].

*System heterogeneity*

During the collaborative training of the global model, different nodes have different capacities (e.g., CPU, Battery, Memory, Bandwidth). As result, if we ignore this fact, the convergence will be very slow, and the weak clients will exhaust their resources before the end of the training, resulting in bad model performance.

*Statistical heterogeneity*

When FedAvg was first proposed by [46], it was based on the assumption of *independent and identically-distributed* (iid) data across nodes, which guarantees a theoretical solution for the equation 4, regarding balanced local data-sets $D_i$, by using the gradient descent optimization method. However, this assumption cannot be held for the majority of distributed data on IoT and users' devices; this is a big limiting factor facing the deployment of Fed ML in real-world scenarios [76]. In fact, the majority of works published on this topic display good results for iid data and poor ones for non-iid setups, which is shown by a bad impact on the global model's performance and the required time and energy for the training [76]. This substantial problem has driven several teams to develop techniques to adapt the original federated learning algorithm to both types of heterogeneity [11, 34, 70].

## 3.2 Energy consumption formulation

The main goal of FedML optimization for energy-constrained devices is to minimize the functional energy consumption of the nodes while building a good global model. In general, a wireless device's total energy consumption ET can be divided into three major parts: Enet, Ec, and Esys (Equation 5).

$$E_T = E_{net} + E_c + E_{sys} \quad (5)$$

$E_{net}$ is the energy consumed by the device for communications with other devices or the server for update exchanges. Ec is the energy consumed by the device's local processing unit and memory to accomplish the training computations. $E_{sys}$ is the energy consumed by the general system operations of the device, which are not related to its participation in the Federated training.

Note that $E_{sys}$ is generally small and negligible compared to the total amount used in IoT [45]. In addition, it is not specific to the problems considered in this study, so we omit it from this formulation.

Moreover, communications generally consume more energy than processing, for an equivalent amount of operations (this justifies multiple aggregation approaches before data transmission). Equation 6 gives the amount of energy consumed by network communication, expressed by a set of parameters related to our context.

$$E_{net} \simeq \sum_{i=1}^{N_T} N^i_{Tbit} \frac{P_T}{R_T} + \sum_{i=1}^{N_R} N^i_{Rbit} \frac{P_R}{R_R} + c \quad (6)$$

$N_T$ and $N_R$ are, respectively, the number of transmitted and received updates by the device. $P_T$ and $P_R$ are the transceiver power at transmission and reception, respectively. $R_T$ and $R_R$ are bit rates for transmission and reception, respectively. $N^i_{Tbit}$ and $N^i_{Rbit}$ are the number of bits transmitted and received, respectively, in a given update i, and c is amount of energy consumed by irrelevant factors such as channel noise, transmission errors, etc.

If $P_R = P_T = P$ and $R_R = R_T = R$, the equation 6 can be simplified into equation 7:

$$E_{net} \simeq \left(\sum_{i=1}^{N_T} N_{Tbit}^i + \sum_{i=1}^{N_R} N_{Rbit}^i\right)\frac{P}{R} + c_1 \quad (7)$$

Moreover, the energy consumption by local computations on each client device is approximated by the equation 8.

$$E_c \simeq \sum_{i=1}^{N_{round}} T_{training}^i \times P_c^i \quad (8)$$

Where $P_c^i$ is the consumed power per training time unit at round $i$, $T_{training}$ is the duration of computation operation, and $N_{round}$ is the number of operations to run by a given device.

If $T_{training}$ and $P_c$ are equivalent for all rounds on a given device, the equation 8 can be simplified as:

$$E_c \simeq N_{round} T_{training} P_c \quad (9)$$

In summary, the approximated total energy consumed by each client device (Equation 5) can be expressed by equation 10.

$$E_T \simeq N_{round} T_{training} P_c + \left(\sum_{i=1}^{N_T} N_{Tbit}^i + \sum_{i=1}^{N_R} N_{Rbit}^i\right)\frac{P}{R} \quad (10)$$

From the above energy formulation, we can identify a list of parameters which impact the energy consumption of the participant client devices in Federated Learning: the number of exchanged updates $N_T$ and $N_R$, the number of bits in each exchanged update $N_{Tbit}^i$ and $N_{Rbit}^i$, the transmission power $P$, the transmission bit rates $R$, the duration of local training $T_{training}$, and the number of local training rounds $N_{round}$.

### 3.3 Fed ML optimization parameters

Based on the established equations in the previous section, together with the studied selected papers, we identify a number of energy optimization aspects.

Accordingly, in order to minimize the total energy in equation 9, the optimization of the local training tasks to accelerate the model convergence should result in decreasing the number of federation rounds $N_{round}$. Moreover, the training time duration $T_{training}$ will be improved if we reduce the trained model's complexity, which impacts energy efficiency. Aggregating updates with the least cost, by reducing the size of exchanged data with the central server (*i.e.,* decreasing $N_{Tbit}$ and $N_{Rbit}$ in equation 7), will help save battery life. Furthermore, the frequency of model update exchanges affects the total number of updates $N_T$ and $N_R$ (equation 7), thus optimizing even more the energy consumed in communications. More optimization can also be achieved by making smart use of the heterogeneous nodes' computing resources to participate in the training, in addition to optimizing the client selection to balance the load over the participant nodes and involve the best ones for accelerated convergence. Finally, decreasing the transmission power $P$ and maximizing the bit rates $R$ (equation 7) also helps to reduce the total spent energy.

This analysis will help us later to classify the different approaches and techniques proposed in the literature, as we will see in subsection 5.3.

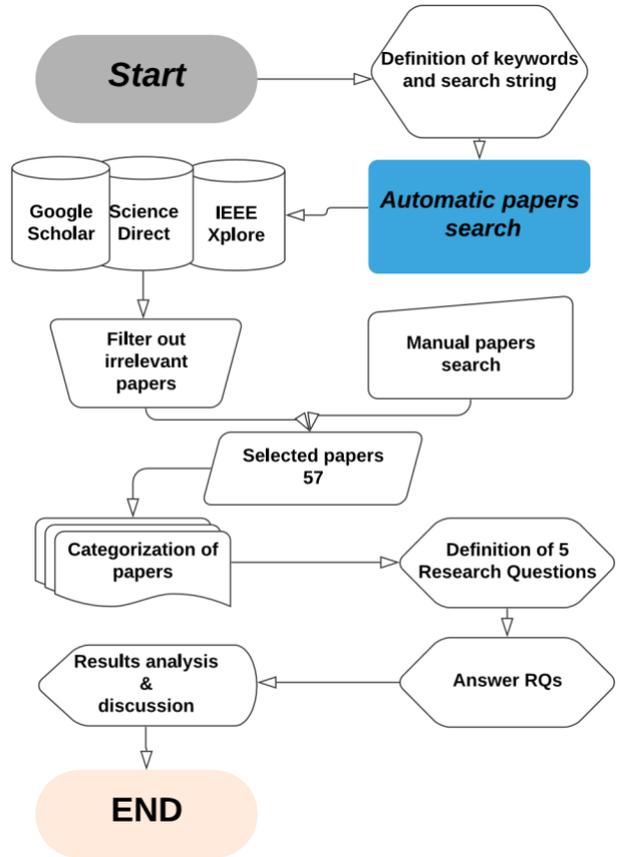

**Fig. 2** Our Search Process

## 4 Systematic Mapping Study Process

This section describes the process followed throughout this Systematic Mapping Study.

Additional material is available on the online repository created for it [1].

Figure 2 illustrates the steps taken. After an automatic search based on the defined keywords and search string in the three common databases, the first step consists of filtering relevant papers based on their title. Then, we refined the selection based on the abstract. We refined our search even further by reading the full text.Finally, we added a manual search step afterwards to spot any articles that were not found the automatic way. Details about each step of the workflow will be presented in the upcoming paragraphs.

## 4.1 Papers selection

In order to obtain all relevant papers for our study, we have queried three main databases (Google Scholar, IEEE Explore, and ScienceDirect) by using the search string in Listing 1, built mainly using the following keywords: federated machine learning, edge computing, on-device intelligence, energy, and optimization.

**Listing 1** search query

*"(" Federated Machine Learning" OR "*
  *Federated Learning") AND ("edge computing" OR*
  *"on–device intelligence ") AND ( energy OR power)*
  *AND ( optimization OR optimal OR efficient OR*
  *efficiency )"*

**Filtering papers**. We filtered the initial search results to keep only papers, that meet all the following inclusion and exclusion criteria.

**i. Inclusion criteria:**

- Papers from 2016 to July 2021
- Papers in the English language
- Papers which propose an optimization of Federated Learning w.r.t. energy consumption, using techniques including communication cost, or training time reduction
- Papers which target the IoT or mobile devices in general

**ii. Exclusion criteria:**

- Works on distributed machine learning with no explicit application to federated learning on resource-limited devices
- Similar works of the same authors

**Manual searching**. In order to cover the literature as much as possible, another step was added to look for potential papers that might have been missed earlier: backward snowballing by looking at cited references in the selected papers. Thereby, additional papers were added for a total of 67 papers. In the remainder of this study, we will refer to selected papers by identifiers, attributed according to the chronological order of the publication: P1, P2, up to P67. The list of all papers, along with their classification, is depicted in Table 5 in Appendix A.

**Table 2** Research questions

| RQ ID | Question |
|---|---|
| RQ1 | What is the publications tendency? |
| RQ2 | What network architectures are proposed? |
| RQ3 | How is the energy optimization achieved? |
| RQ4 | How is the optimization validated? |
| RQ5 | What are the reported optimization results ? |

## 4.2 Research questions

In order to analyze the literature and compare the proposed techniques in a systematic way, we define a set of research questions that will guide our analysis (Table 2). RQ1 indicates the timeline and sources of the papers; RQ2 presents the network topology considered by each paper; RQ3 examines the FedML energy optimization aspects that are addressed by each paper; RQ4 presents the experimentation setups used to validate the approaches; and RQ5 measures the optimization improvements of the experiments.

---

[1] https://gitlab.com/rachid-el-mokadem/fedmlsysrev

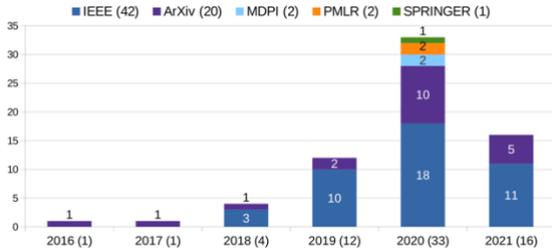

**Fig. 3** Fed ML papers publication trend over time

# 5 Questions answering

In this section, we present the results analysis from the study of the selected papers, arranged as answers to the research questions defined in subsection 4.2.

## 5.1 RQ1 - What is the publications tendency

Answering this research question will account for providing the number of publications evolution, their distribution over the publishing venues, and the nature of papers, as well as their influence on the field of FedML.

The graph in Figure 3 shows the papers publication trend over time. The growing number of papers over the last 3 years is clear, with 33 papers in only 2020. Given that the first paper from [32] was published in 2016, we can clearly see the big interest this subject is receiving from several research teams around the world.

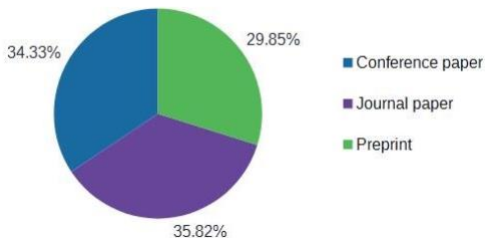

**Fig. 4** Paper types distribution

The majority of papers, as shown in Figure 4, were published in journals (≈36%) and conferences(≈34%). This shows the growing maturity of this subject and the engaged efforts by the scientific community. We also have 20 out of 67 (≈29%) papers published as pre-prints on the ArXiv database, including 10 in 2020. This could be justified by the fact that the subject is evolving quickly, with fast feedbacks. We have also included the non-peer reviewed papers of Konecny´, Jakub et al. [32, 33], since they are considered the most impactful in the subject, with 747 and 1733 citations, respectively. The same team is behind the seminal work on the FedML proposal [46].

Furthermore, we consider the number of citations for each paper, shown in Figure 5, to measure their influence on the subject. It is obvious that older papers tend to get more citations than new ones. However, it does provide an approximate idea of the paper's scientific interest for the community. From the graph, we notice some spikes on a couple papers. For older papers such as P1 through P8, this is somehow reasonable. However, in the case of P15 ([57]) with 345 citations, P19 ([8]) with 110 citations and P35 ([54]) with 145 citations, this definitely shows the high impact of those papers. More details on the techniques used by them in subsection 5.3

## 5.2 RQ2 - What network architectures are proposed

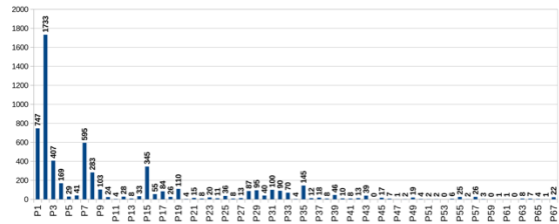

**Fig. 5** Number of citations per paper

In this question we consider the proposed network architectures of the studied papers. This is important to us, because the network topology has an impact on the communication cost, and therefore the energy consumption.

The architectures are as follows.

- **Centralized**: based on a central server to ensure the communication and model's parameters exchange, between the participating devices. This option is energy consuming, due to long range communication between the devices and the server, which requires higher transmission power $P$ (equation 7). It also suffers from a single point of failure.

- **Decentralized**: based on node to node communication without the need for a central server. In this setup, the devices can save lot of energy, by opting for short range communication between the nodes only [15].
- **Hybrid**: this architecture is based on at least three layers of devices, where intermediate ones are placed between the central server and the end devices.

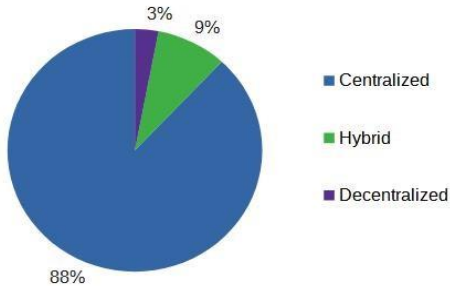

**Fig. 6** Number of papers by Network topology

The hybrid architecture is based on adding edge servers between the main server and the end devices. These intermediate devices can play several roles, such as managing direct clients under their control, which allows the offloading of the central server and lowers the waiting time for aggregating multiple received updates. In some cases, this edge server can also be used to offload the end devices from local update computing, by periodically querying the training data from the selected clients, doing the updates with a much higher computing capacity and communicating with the server, on behalf of the end nodes. As a result, this architecture can allow a high energy optimization on the devices, although posing some threats to data privacy, especially when these edge servers are not trustworthy, and the data is very sensitive.

Figure 6 shows that the majority of papers (59 out of 67) are based on a centralized setup, while 2 papers have a fully decentralized one, and 6 propose a hybrid architecture. The predominance of the centralized scheme can be explained by the influence of the architecture in the original paper [46], which comes from Google. Moreover, the fully decentralized scheme faces some algorithmic and practical challenges to aggregate the models without a central device [30].

## 5.3 RQ3 - How is the energy optimization achieved

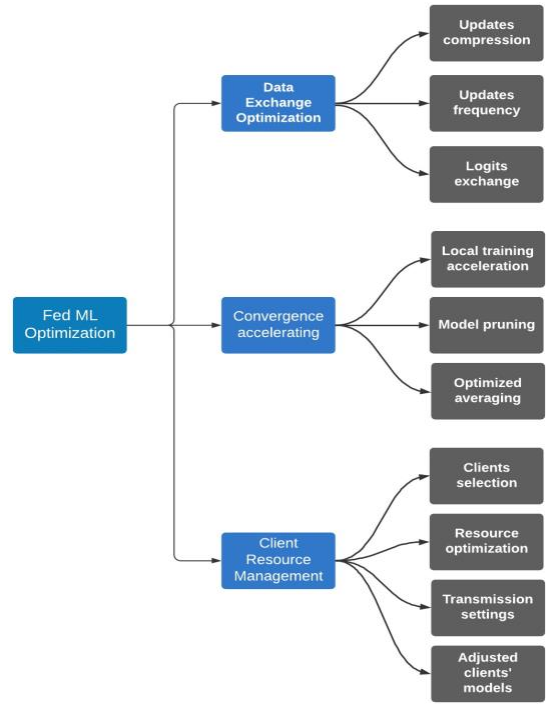

**Fig. 7** Federated Learning optimization techniques (recap)

In this question, we analyze the techniques used by the papers, to optimize the Fed ML. Our study focuses on the power consumption reduction, so as seen in subsection 3.2, all studied optimization aspects are linked with the energy through equation 10. We classified these techniques into the following categories: (1) convergence acceleration (2) data exchange optimization and (3) client resource management. Figure 7 recaps the different techniques. Table 3 presents the optimization aspects addressed by each studied paper.

### 5.3.1 convergence acceleration

In federated machine learning, the training tasks are performed by the client nodes to build a global model under the orchestration of the central server, during as many rounds as needed to reach a good performance. In order to save the battery life of the client devices, the total time to reach global model convergence can be reduced with several approaches.

*Local training acceleration*

Many works have used different optimizations to accelerate the local training, such as adaptive learning rate [42], and Adam optimization method [47].

Equation 11 is used in the original version of Federated Learning. $w$ are the model weights, $b$ is the model bias, $\nabla$ is the gradient of the loss function $l$ and $\mu$ is the learning rate.

In this version, the server defines a learning rate parameter at the beginning, used to compute the gradient descent steps in local training. Opposed to that are the aforementioned methods, which determine the best steps to take in order to quickly achieve the convergence of the global model.

$$w \leftarrow w - \mu \nabla l(w, b) \quad (11)$$

The benefit of these techniques is to decrease the number of rounds $N_{round}$ (equation 9) required for the model convergence, and thus reduce the energy consumption for the participant devices.

Accordingly, [47] proposed CE-FedAvg, which improved how the nodes compute their local updates by using the Adam method, known for its improved learning rate, instead of SGD (used in the original FedAvg algorithm). The weights' update method of the proposed algorithm, executed by each client, is shown in equation 12.

$$w_k, m_k, v_k \leftarrow AdamSGD(w_k, m_k, v_k) \quad (12)$$

Where $w_k$ are the model weights, $m_k$ is Adam's first moment, and $v_k$ is Adam's second moment. These parameters are used to compute the Adam steps by averaging them over all received updates or gradients and sending them back to the clients in the next round of the training.

Feature augmentation is a technique used in machine learning to improve training performance in an unbalanced class distribution ([75]). Similarly, in the context of Federated Learning, FedFusion is an algorithm presented by [72] to accelerate the global model's training by using a technique named Feature Fusion, which is based on using a combination of the global model's feature space with the local model's feature space to train the local model. The global model is used as a feature extractor, and then multiple types of feature fusions are employed to efficiently aggregate all of them. Additionally, [71] presented a two-Stream model learning with Maximum Mean Discrepancy (MMD), where the nodes training is performed on two models, in parallel, both initialized with the global model parameters, but one of them (global model) is kept unchanged during the training. An MMD loss is computed between the output of the two models, which is used to optimize the local one. This technique is often employed with learning transfer and knowledge distillation in standard machine learning, and its adoption for Federated Learning helps to accelerate the training and reduce the communication cost. In essence, it consists in constraining the local model training by the global model parameters, to avoid that local models over-fit their local data, thereby building a good global model in lesser training rounds $N_{round}$. [4] used an adaptive dropout schema to decrease the convergence time by reducing the local model's complexity and number of trainable parameters. In practice, each round a random sub-net $w_c$ of the global model is sent to each participant client $c$, then an activation score map $M$ is used to track the indexes $A$ of the best sub-models to be reused in the next rounds.

*Model pruning*

Model pruning is another technique widely used in deep learning, which accelerates the training, by reducing the number of model parameters, based on training data. The reduction simplifies the model, thereby decreasing the computation time ($T_{training}$ in equation 9), local training energy consumption $E_c$, while keeping a good model performance. [29] implemented an algorithm named PruneFL where the pruning is performed initially by a selected client on its local data. Then the resulting smaller model is iteratively adapted by the server in each round w.r.t. to the training efficiency, by involving all clients updates, to reconfigure it, through removing or adding back some parameters. In order to allow the reversibility of parameters adding and deleting, the authors used a mask with zeros and ones for removed and kept weights respectively. Similarly, [69] proposed a structured model pruning combined with weights quantization and selective update, to accelerate the training and reduce the computation cost on the devices. In particular, the authors used an $l_1 - norm$ based pruning of the model weights with a variable ratio from 0 to 90%.

*Optimized averaging*

While original Federated Learning works by gathering the local model updates, and simply averaging them, several papers proposed to use advanced averaging methods, allowing a fast training convergence. Accordingly, [23] proposed Federated Momentum (FedMom), a technique with biased gradients that uses the momentum method to update the global model, according to equations 13 and 14:

$$v_{t+1} = w_t - \eta \sum_{k=1}^{K} \frac{n_k}{n}(w_t - w_{t+1}^k) \quad (13)$$

$$w_{t+1} = v_{t+1} + \beta(v_{t+1} - v_t) \quad (14)$$

Where $v_t$ is the average of the previous round's updates and beta $\beta$ (often equal to 0.9) is the parameter used to compute the moving average of the updates, through time. On the other hand, [39] used a hierarchical architecture by introducing $L$ edge servers between the central server and client nodes. Each edge server has a subset $s$ of clients from which it aggregates the updates before forwarding them to the main server. According to the authors, this method reduces training time and decreases node energy consumption.

### 5.3.2 Data exchange optimization

The global model is built by gathering and aggregating the updates from the participant nodes at the central server. The frequency of exchanging the computed updates and their data sizes are optimized by several works in order to achieve communication-efficient federated learning, which drastically saves the battery life of the participant nodes without compromising the global model's performance. In FedAvg, the aggregation of the local models is achieved according to the following equation 15:

$$w_{t+1} \leftarrow \sum_{k}^{K} \frac{n_k}{n} w_k^{t+1} \quad (15)$$

Where $w_k$ are the learned weights at each node, $n_k$ are the number of data points at each node, and $n$ the total number of data points for all $K$ participant clients.

*Updates compression*

A stated limitation of FedAvg [46] is that the participant clients must upload the full computed updates at each round of the training, which has an impact on the power consumption of these devices. The proposed optimization methods allow the reduction of the data exchanged between the nodes and the server while preserving the quality of the global model.

In addition to ordinary data compression algorithms used to encode the final updates with lower amounts of bits, such as *Huffman encoding* used by [7, 43], data size reduction is achieved by several other methods, such as update *quantization*, *sparsification*, and *sketching*. The goal of all these techniques is to reduce the amount of bits per round $N_{Tbit}$, sent through the wireless interface, which subsequently decreases significantly the energy usage for exchanging model's updates (equation 7).

The *quantization* of machine learning models is based on using low float-point precision to represent the model's weights in order to reduce the bit size ([2]). [68]proposed a method called Federated Trained Ternary Quantization (FTTQ), which reduces both upstream and downstream traffic. It implements a layer-wise weight quantization with an adjustable threshold during the training, which has the additional benefit of reducing the training tasks' energy budget. Similarly, [27, 43, 47, 54] used quantization for data size reduction, in most cases mixed with other techniques. Furthermore, [27, 44] proposed an adaptive schema for updating quantization to achieve communication-efficient training.

Additionally, the *sparsification* of the global and local models is used to compress the exchanged data by eliminating the gradient values of the computed update that are below a given threshold and replacing them with zeros. This operation results in a sparse model update that can be encoded with a small number of bits in order to optimize the communication cost and energy budget. Accordingly, [57] proposed Sparse Ternary Compression (STC), a new compression framework created especially for the requirements of Federated Learning on resource-limited devices. STC extends the existing compression methods (in particular top-k sparsification [60]) to support downstream compression; additionally, the authors combined sparsification with quantization and Golomb encoding to achieve better optimization results. [20] developed an *adaptive gradient sparsification* based on bidirectional top-k gradient sparsification to reduce communication costs in both directions between the server and the nodes. The sparsification's parameter $k$ is determined by the server as a trade-off between communication and global model accuracy. Moreover, [62] used a gradient sparsification with gradient correction, in order to accumulate the insignificant eliminated gradients and add them lately to speed up the convergence of the model training.

Other techniques used to this end are gradient *sketching* [33, 55] and subsampling. The first one is based on compressing the update with a data structure named Count Sketch [61]. The second one [33] involves clients, sending only a smaller update derived from the computed one, by randomly sampling their values. The server then averages all received sub-sampled updates to get an estimate of the global model parameters. Additionally, [52] used dual-side low-rank compression to reduce the size of the models in both directions between the server and the nodes. Finally, [36] used a layer-based parameter selection in order to transfer only the important parameters of each model's layer.

*Updates frequency*

In the original Fed ML algorithm, clients send updates at each iteration of model training, which induces high communication costs and energy consumption as the number of updates exchanged with the server $N_T$ and $N_R$ increases. In order to perform, under a given resource budget, [65] proposed a control algorithm that determines the best trade-off between local update and global parameter aggregation. It learns the data distribution, and system characteristics along the distributed training, then determines dynamically the frequency of global model aggregation, with respect to the resource constraints. Alternatively, [8] presented a different method, which is based on the model's layer-wise frequency. It means that important layers' parameters are more frequently exchanged than less important ones. The reason is that the first layers of a deep model tend to learn general features for different data sets, while the deeper layers learn more particular ones. Consequently, each node separates its model into shallow layers' weights $w_g$ and deep layers' weights $w_s$, which are exchanged with the server separately and asynchronously, under the control of the server. It determines, for each client, the type of weights to consider, and performs a temporally weighted aggregation to give more importance to the newest received models. [64] proposed another method where the clients pull the global model less frequently from the server (to reduce down-link energy consumption) and compensate the gap with local updates.

*Logits exchange*

Some works have chosen not to exchange the updates with the server: only the outputs of the trained local models, called *logits*, are sent to the server which reduces drastically the communication cost of the federation by many orders of magnitude [58]; nevertheless, all clients and the server must have shared public data samples to compute and share their outputs.

In order to build a global model out of this reduced data, the authors of [24, 26, 50, 58] used a learning transfer technique called *knowledge distillation* [59], where multiple teacher models (local models) transfer their learning to a single student model (the global model) [21]; In all cases, the distillation task is performed by the server, except for [24] where the sent logits are averaged by the server and sent back to the clients to perform the distillation themselves. In that case, the communication cost is reduced in both directions as the server does not send the whole model to the nodes.

### 5.3.3 Clients resource management

Many approaches allow client devices to participate in model training, with optimal energy. Client resources generally refer to CPU time, memory and wireless bandwidth, which are often related to energy consumption. Two of the most used approaches for client resource management affect (1) client participation and (2) transmission settings.

*Clients selection*

In the original FedML, participants were selected randomly, each round, from the available nodes. Subsequent analysis showed that this approach yields poor model convergence and causes node resources to be wasted ([49]). Accordingly, many works have addressed this aspect, by *adaptive* and optimal client selection, based on their available resources and data in each round. [49] presented FedCS, a Federated Learning algorithm with optimized client selection, where the server starts by selecting a random set of clients, then performs a more informed selection using client resources and the time taken to compute the updates. Furthermore, [3] presented a *Reinforcement Learning* scheme at the server, based on energy units $e_n$, number of CPU cycles $c_n$, and the amount of data points used for training by each client, per-round.

A server reward is then computed from these values to help it find the best policies and actions for efficient training with optimal resource usage. In the same vein, [53] proposed a multicriteria client selection model, named FedMCCS, that is based on a discriminative selection of client devices based on CPU, Energy, Memory and Time. The server tracks these values along the training by an auxiliary data exchange of requests/responses with the nodes. A linear regression model is trained on these attributes to predict whether a client has enough resources to participate in training tasks. Moreover, [67] proposed the selection of clients based on their participation history, which impacts the global model's performance. Additionally, [56] proposed a data imbalance aware selection of the participants in each round, such that all data categories must be covered at least once. This is achieved by requesting a bit-mask $\eta$ containing $C$ bits corresponding to the available data categories from each node. The server then sorts these bit masks in a decreasing order of the number of sets and minimizes the required number of clients to get all categories covered by the averaged updates.

*Hybrid scheme*

Other papers have proposed a *hybrid scheme*, based on the architectures presented in our second research question, to optimize the resources of the client devices. [12] developed a self balancing system based on mediator edge servers, gathering near uniform data distribution subsets of clients, and aggregating the trained models, before sending them to the central server to build a global one; Similarly, [39] achieved energy consumption reduction by balancing the exchange of parameters with $L$ edge servers with respect to the training time and communication budget where each edge server incorporates a small number of clients [66] used a hierarchical aggregation of the model updates to overcome the communication overload between the nodes and the server. Moreover, [25] and [74] proposed a cloud-edge-client scheme wherein the clients offload a part or all the training tasks to the edge servers, which get portions of the clients' data for the training. This approach has some flaws w.r.t. communication overhead and privacy concerns for clients' data, but it may be relevant in some application-specific scenarios.

*Transmission settings*

Wireless transmission has a high energy cost for mobile and IoT devices in general. As we saw in equation 7, it is related to the transmission power $P$ and the bit rate $R$ of the network interface. Many papers have tackled the optimization of wireless transmission and bandwidth settings. [48] presented a technique for transmission power and rate optimization for energy-efficient communication between the nodes and the central server. They consider minimizing total energy consumption as a joint optimization problem over bit rate, transmission power, and CPU frequency for local updates. On the other hand, [28] proposed a fully decentralized communication between the nodes, based on a Segmented Gossip protocol. In this communication scheme, the workers segment their updates into small fragments that are separately shared with a subset of the participants, who then relay them progressively to other workers until all of them get everything. The peer workers with faster bandwidth are chosen to pull updates from. Moreover, [63] presented an energy-aware worker scheduling algorithm: a node monitors its energy consumption at each round and, by comparing it with an adjustable threshold, decides whether to participate in the next training round or not. Subsequently, each worker partitions its update into M segments, which are transmitted separately on M sub-channels with adjusted transmission power.

*Adaptive local models*

Recently, [73] and [11] proposed to adapt the local models to the nodes' capabilities and data, in opposition to the state-of-the-art FedAvg algorithm, where all clients get the same model architecture. In both works, an adapted model is derived for each client as a sub-net of the global model, with a smaller number of layers and parameters. Then, different averaging methods are used to aggregate the updates. These new methods are heterogeneity and data imbalance aware and allow an adaptive saving of energy used for training and communications.

From Figure 8, we see that the d*ata exchange optimization* category has the highest number of papers (32), followed by *client resource management* with 25 papers, and finally *convergence acceleration* (10 papers). Figure 9 shows in more detail the optimization techniques used by each paper. It was not surprising that Updates Compression and Clients selection represented most of the papers because the original FedAvg algorithm had a substantial limitation on these aspects. Additionally, this will have the highest impact on devices' energy preservation, knowing that wireless communications use the biggest part of the operational energy. The other techniques are shared between the rest of the papers, with a small advantage for Local training acceleration.

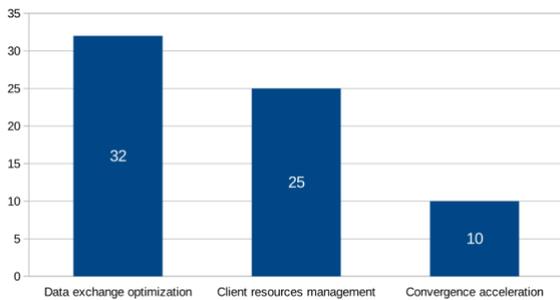

**Fig. 8** Papers count per optimization category

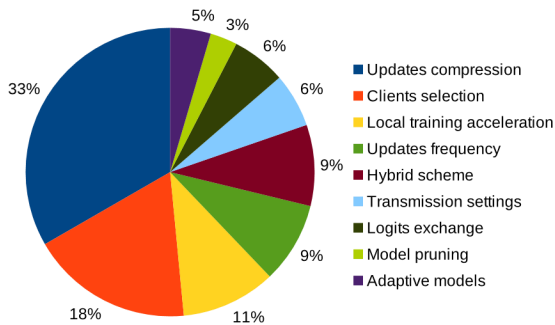

**Fig. 9** Papers count per optimization technique

## 5.4 RQ4 - How is the optimization validated

This Research Question is about the experimentation setup (models, data sets, and testing platforms) used by different papers to validate their proposed works.

The classification data in Table 4 shows that the majority of papers considered only neural networks (specifically convolutional ones) with MNIST and CIFAR10 data sets for the experimental part of their research. While this seems to restrict validation, it can be explained by the ease of getting these famous data sets and implementing NNs on top of popular machine learning frameworks such as PyTorch and Tensorflow. Some papers have implemented additional validations on other ML models such as Linear Regression, Logistic Regression, and Support Vector Machines. However, there is a substantial shortage of validation results for non-neural network models, which may exhibit lower complexity and therefore lower resource consumption.

As for the experimentation platforms, different papers considered different numbers of participating nodes, from 2 up to 50000. The majority of papers (52 out of 57) have used emulated nodes on multi-GPU computers, while only five papers (P7, P14, P17, P34, P58) have performed experiments on real devices such as the Raspberry Pi™and smartphones. Emulated devices can give an insight into validation, but it is important to have further results on real ones, in real life scenarios, especially regarding wireless communications, energy constraints, and computing power. The democratization of rapid prototyping platforms in the industry and academia (*e.g.,* Arduino™, Raspberry Pi™ & ESP31™) is another motivation for that.

## 5.5 RQ5 - What are the reported optimization results

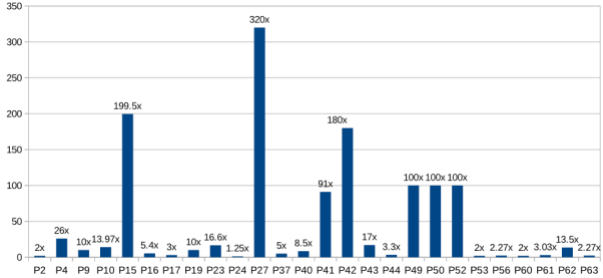

**Fig. 10** Communication cost improvements (per paper)

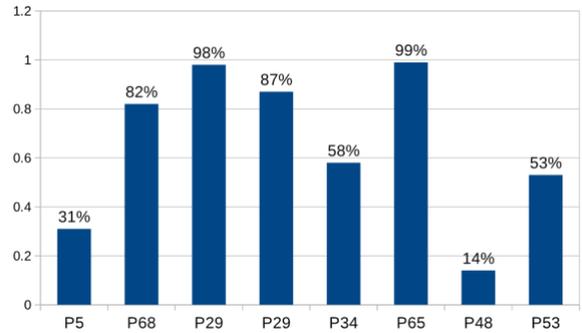

**Fig. 11** Energy consumption improvements (per paper)

**Table 3** Papers list by optimization techniques

| Category | Optimization technique | Papers |
| --- | --- | --- |
| *Data exchange optimization* | Updates compression | P2 P15 P17 P19 P23 P45 P24 P26 P27 P35 P37 P39 P40 P42 P64 P52 |
| | Updates frequency | P1 P7 P8 P20 P56 |
| | Logits exchange | P4 P10 P46 P49 |
| *Client resource management* | Clients selection | P3 P5 P18 P25 P28 P31 P32 P38 P44 P47 P48 |
| | Hybrid scheme | P16 P29 P34 P36 P51 |
| | Transmission settings | P67 P22 P30 P57 |
| | Adaptive models | P54 P55 |
| *Convergence acceleration* | Local training acceleration | P6 P11 P12 P13 P33 P63 |
| | Model pruning | P14 P58 |
| | Optimized model averaging | P21 |

In this question, we list and compare the optimization results obtained in the surveyed papers. The numerical results are classified into three categories: (1) Communication Cost, (2) Convergence Time, (3) Energy Consumption. Each paper quantified its optimization improvement compared to the standard FedAvg [46] algorithm, and the numerical results w.r.t. each category are listed in the graphs. From Figure 10, we see a very wide range of improvement values related to global communication cost reduction, which goes from 2x up to 320x. In Figure 12, we see convergence time improvements going from 3% up to 98%. As for the energy consumption, Figure 11 reports a range of improvement from 14% to 99%.

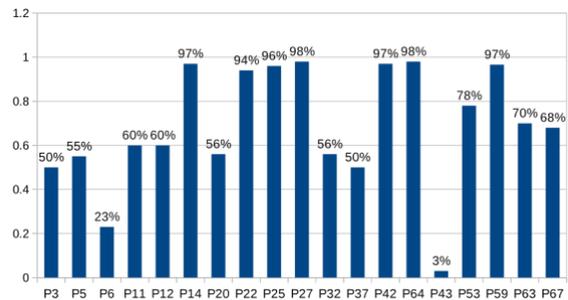

**Fig. 12** Convergence time improvements (per paper)

The convergence time and communication cost optimization results are very encouraging, which is consistent with the important interest of the community in these two aspects (Figure 8). Although they directly impact the energy consumption, note, that during our readings, relatively few papers (8 out of 67) have evaluated the optimization's benefit *directly* on the energy consumption, which is crucial to our study.

# 6 Discussion

In this section we discuss the Systematic Mapping Study results, in the light of the previous analysis, guided by the RQs in Section 5. For each *Research Issue* (RI) presented, we provide (1) some remarkable limitations related to Energy constrained Fed ML, and (2) some improvement directions and recommendations for the research community.

## 6.1 RI1: Fully-decentralized scheme

In a centralized scheme based Fed ML, client nodes exchange data during the training with a central server, which is generally located in the cloud. Consequently, the nodes have to use long-range wireless communication to reach the server, which implies high power consumption [15]. To overcome this, we must take advantage of the short-range communication between the nodes, which is by far less power-intensive, to exchange the updates using peer-to-peer communications.

The proposed approaches to implementing a fully decentralized FedML induce an overload on the resource-limited devices, caused by the additional operations performed by the nodes to compensate for the role of the central server. In [28], the nodes have to also play the role of the aggregating server, and [13] proposed a technique where all nodes compute and exchange their updates in a chain-like scheme (using some sort of multi-hop

**Table 4** Papers validation setups

| Paper | ML model | Dataset | Number of nodes |
|---|---|---|---|
| P1 | RNN | Blog posts dataset | 1000 |
| P2 | CNN - RNN | CIFAR10 - public post reddit | 100 - 1024 |
| P3 | CNN | CIFAR10 FashionMNIST | 1000 |
| P4 | CNN | MNIST | 10 |
| P5 | RNN | - | 3 |
| P6 | CNN : AlexNet | CIFAR10 - MNIST | 2 – 100 |
| P7 | CNN | CIFAR10 - MNIST | 5 (RaspberryPi) – 500 |
| P8 | - | - | 50 |
| P9 | CNN - LSTM | MNIST | 30 |
| P10 | ANN | MNIST | 25 |
| P11 | CNN | MNIST - CIFAR10 | - |
| P12 | CNN | MNIST - CIFAR10 | - |
| P13 | CNN | CIFAR10 | 20 |
| P14 | CNN: VGG11 - LeNet | FeMNIST | 5 - 10 (RaspberryPi) |
| P15 | CNN: VGG11 - LSTM - Logistic Regression | CIFAR10 - KWS dataset - FashionMNIST - MNIST | 100 |
| P16 | CNN | EMNIST - CIFAR10 - CINIC10 | 500 |
| P17 | CNN | MNIST CIFAR10 | 10 - 40 (RaspberryPi) |
| P18 | Linear model | Random integer data | 20 |
| P19 | CNN RNN | MNIST HAR | 20 |
| P20 | ANN | MNIST | 15 |
| P21 | Logistic Reg - CNN (ResNet18) | CIFAR10 | 20 |
| P22 | CNN | CIFAR10 - MNIST - SVHN | 20 |
| P23 | CNN | FEMNIST | 50 |
| P25 | Regression model - CNN - SVM | Boston Housing dataset - MNIST - KDD Cup'99 dataset | 5 - 100 - 500 |
| P26 | SNN | MNIST-DVS dataset | 2 |
| P27 | CNN (LeNet-5 – CifarNet – DenseNet-121) | MNIST - CIFAR10 - ImageNet | 64 |
| P28 | CNN | MNIST | 50 |
| P29 | CNN | MNIST - CIFAR10 | 50 |
| P30 | ANN | MNIST | 50 |
| P31 | CNN | - | 80 |
| P32 | CNN | Fashion/MNIST - CIFAR10 | 100 |
| P33 | CNN | MNIST | 3 |
| P35 | Linear Regression - CNN | MNIST - CIFAR10 | 50 |

| P36 | CNN | MNIST | 500 Clients/10 Edge server |
|---|---|---|---|
| P37 | CNN - Linear Regression | MINST - California Housing dataset | 10 |
| P39 | CNN (AlexNet) - Transformer (GPT2-small) | CIFAR10 - PersonaChat | 10000 to 50000 |
| P40 | CNN - Logistic Reg | CIFAR10 - FashionMNIST - Sentiment140 | 100 |
| P41 | CNN | CiFar10 - MINST | 10 |
| P42 | CNN MNIST | CIFAR10 | 1 to 64 |
| P43 | CNN (ResNet) | CiFar10 - CiFar100 | 16 |
| P44 | ANN | MNIST - FEMNIST | 100 |
| P46 | CNN | MNIST - CIFAR10 | 10 to 20 |
| P47 | CNN | FeMNIST - CiFar10 - CiFar100 | 100 |
| P49 | CNN - LSTM | MNIST - IMDb | NA |
| P50 | CNN (ResNet) | CIFAR10 | 10 |
| P52 | CNN | MNIST - EMNIST | 20 |
| P53 | ANN - CNN | EMNIST | 50 - 1000 |
| P55 | CNN (ResNet) - Transformer | MNIST - CIFAR10 - WikiText-2 | 100 |
| P56 | CNN | MNIST | 30 |
| P57 | CNN | MNIST - CIFAR10 and SVHN | 20 |
| P58 | CNN | MNIST - CIFAR10 | 3 (Core i5 PCs) |
| P59 | CNN | MNIST | 10-18 |
| P60 | CNN | CiFar10 - FEMNIST - IMDB | 10 |
| P61 | CNN | CiFar10 | 10 |
| P63 | CNN | FEMNIST - Shakespare - Sentiment140 | - |
| P64 | CNN | MNIST – Cifar10 | 12 |
| P65 | CNN | CiFar10 - FashionMNIST | 8 |
| P66 | CNN | MNIST - FashionMNIST - CIFAR10 | - |
| P67 | CNN - LSTM | FEMNIST - Shakespeare dataset | - |

communications). The limitation of the first technique is the overhead tasks for the nodes to play the server role, where the second one forces the nodes to run all the time of the training: in both cases, more energy and resources are required at the node level

The hybrid scheme seems to overcome some of these problems since it has a cloud-based central server that is only used to manage the client participation and selection, with very limited data querying from the nodes, while keeping model aggregation between the client devices. At the same time, in order to advocate the fullydecentralized scheme, there is a need for a new theoretical framework that supports complete decentralized model aggregation with convenient energy and resource consumption.

## 6.2 RI2: Large models reduction

Some Fed ML models with large sizes and a big number of trainable parameters (*e.g.,* Deep Neural Networks) require a computationally expensive training [17] for energy constrained IoT devices.

If we consider Fed ML as a bootstrap aggregation [22] of the global model over different distributed nodes' data sets, we could reduce the local model's size and complexity to make the training tasks easier for the devices. We could still build a high-performance global model by aggregating (*e.g.,* by majority voting) the distributed models repeatedly.

Another possible solution lies in the *Lottery winning ticket* hypothesis, elaborated by Frankle and Carbin [16], which states that a dense neural network contains a sparse sub network, that can be trained to equivalent performance of the initial network. By applying this technique in the context of Federated Learning, the global model can be drastically reduced in size and complexity, to accelerate the training, and reduce the computation load over the nodes. The target sparse model could be obtained from the global model by Adaptive Iterative Pruning (AIP) [19] or Neural Architecture Search (NAS) [51], performed adaptively by the server based on model performance and available client resources.

## 6.3 RI3: Energy-aware data compression

Many papers proposed different techniques to reduce the amount of data to be sent or received from the server [33, 54, 57]. The compression techniques used are: *sketching*, *sparsification*, *quantization*, and data encoding. They have helped to drastically reduce the communication costs for the client devices. However, they add overhead tasks to the nodes, resulting in memory and CPU usage to compress, encode, and decode the transmitted data.

Many works [37, 37] have revealed the benefit of error-controlled lossy compression schemes on the compression rate and computation efficiency. We recommend studying an equivalent technique adapted to Federated Learning's update compression in order to further reduce the communication energy cost.

## 6.4 RI4: Heterogeneity aware optimization

Nodes heterogeneity is a crucial issue for Federated Learning in real world applications [35]. As a result, this topic has drawn the attention of the research community through several works [49, 70], which proposed different methods based on discriminative participant nodes selection. They only choose the devices that have *both* the required resources and data for the training. While this seems to solve the heterogeneity issue, it may impact the model performance and convergence time, by eliminating some devices with either one of those.

In this case, it would be more profitable to manage the devices in a way that takes advantage of their data and computation capabilities separately. Some nodes may participate with their data, others with their computing capacity, and the rest with both. To achieve this flexibility, some devices may exchange raw data, extracted features, or data labels. To preserve data privacy (one of the Fed ML rationales) during these communications, we may use a lightweight encryption technique (*e.g.,* based on elliptic curves) for node-node communication or an homomorphic encryption scheme [18] for untrustworthy node-to-node relationships.

We could also suggest an approach to managing heterogeneity that would be based on the separation of client nodes into two groups. The first one would contain powerful, resourceful devices dedicated to training tasks, and the other one would contain poor nodes for validation only on their own data. The validation score may be used as feedback to adaptively adjust model aggregation parameters.

## 6.5 RI5: Results validation

The majority of studied papers have validated their approach using emulated nodes on powerful computers. Moreover, a substantial focus was given to image recognition tasks using Convolutional Neural Networks (CNNs) and common data sets such as MNIST and CIFAR10. The choice of image data for validation can be explained by its sensitivity and significant size, resulting in elevated communication costs thereby justifying Fed ML usage. However, in IoT and also on mobile devices, there are other types of potential applications with different forms of data and learning tasks (*e.g.,* environmental quantities such as temperature or humidity).

In order to validate the proposed techniques and achieved results in a transparent and replicable way, we underline the importance of conducting an advanced testbed under real-world conditions with real IoT or mobile devices and diverse learning tasks. Moreover, we recommend to build a standardized benchmark for Federated Learning performance analysis, in order to allow researchers from all over the world to validate their works with real diverse data and real-life scenarios.

## 6.6 RI6: Federated inference

The majority of the literature on collaborative machine learning concentrated on the training phase. Although the inference task is less expensive in terms of energy and resources, we may need to consider it in the FedML context with energy-constrained IoT devices to collaboratively compute predictions or classify events. This would also be beneficial in the case of audio and video processing, which involve large amounts of data and models. Moreover, the importance of this topic is apparent in the case where the correlation between multiple nodes is required to classify or predict a value.

In this way, we recommend working on a collaborative inference framework for FedML that allows the nodes to support each other to balance the prediction or classification load instead of relying on the server for this task. Again, appropriate encryption mechanisms have to be used to guarantee data privacy.

# 7 Threats to validity

Any survey or systematic mapping study (including ours) is likely to have some common limitations [10], related to literature coverage and biases in processing the studied items. In order to reduce these threats as much as possible, we tried to follow a well-defined process [31]. It started with a thorough search of relevant papers in different databases, leveraging search term synonyms to get as many valid results as possible. We manually filtered the papers in multiple stages: using the title and the keywords, then reading the abstract, and finally studying the full text. We have repeated this process at least two times: at the beginning, and after a couple of months.

However, since we worked with the resources available at the time, there may be issues related to *search string choice*, the *data collection process*, *research question choice*, and *time span*.

Regarding the search string choice, although we used clear keywords, there may be some missed opportunities due to bad keyword indexation. We have done a manual snowballing from the earlier validated papers, which helped us spot some missed articles by the automatic search process. However, this might not be always enough.

Regarding the *data collection process*, each article was reviewed (title, abstract, and full text) by a single researcher, which might cause some errors. This problem was partially solved by discussions between us.

In relation to the choice of research questions, despite our extensive discussions to be as comprehensive and clear as possible, there could be some aspects that were not covered.

Regarding the time span, we covered the period starting from the seminal paper's publication in 2016 until July 2021. Some interesting papers may have been published after.

Finally, we hope to have more resources in the future to address the previous eventual shortcomings as well as others that our fellow researchers will kindly point out.

# 8 Conclusions and future works

**Summary**. In this paper, we presented the first Systematic Mapping Study, to the best of our knowledge, on Fed ML for Energy Constrained IoT devices. Through a reproducible Research Process, we selected 67 papers related to the topic since the publication of the founding paper by [46] and tried to compensate for eventual biases by snowballing and manual searches.

The results analysis was structured around 5 Research Questions related to publications overall tendency, Fed ML network architecture, and energy optimization schemes (reported results and validation). It appears that updates compression and clients selection have had the highest focus in the literature and yield interesting results in terms of decreasing the communication cost (up to 320x), convergence time (up to 98%) ; and energy consumption (up to 99%).

From our analysis, we identified 6 Research Issues with associated recommendations: fully decentralized schemes, large model reduction, energy-aware data compression, heterogeneity exploitation, real-world results validation, and federated inference. Recommendations include methods, such as, global model size reduction and efficient data compression schemes, to help reduce the communication and computation costs for the nodes. To efficiently address the system heterogeneity, we pointed towards an adaptive and flexible management of the resource-limited devices and involved them in the training. Finally, we underline the need for a standard benchmark, dedicated to a transparent and rigorous validation of the results, with real world conditions and real test-beds.

**Future works**. We plan to conduct a Systematic Literature Review (*SLR*) on the specific topic of fully decentralized Fed ML, which appears to be very interesting. Indeed, it eliminates the single point of failure and presents difficult challenges related to aggregating updates without any focal point. An SLR is dedicated to going in depth regarding a specific question, as opposed to an SMS, which broadly structures the field. Therefore, it is, in our opinion, the logical extension of our work.

# A. Appendix

This appendix lists all papers included in our study, tagged from P1 to P67 (chronological order).

**Table 5** Papers list

| Id | Paper title | Optimization technique |
|---|---|---|
| P1 | Federated Optimization:Distributed Machine Learning for On-Device Intelligence | Data exchange optimization |
| P2 | Federated Learning: Strategies For Improving Communication Efficiency | Data exchange optimization |
| P3 | Client Selection for Federated Learning with Heterogeneous Resources in Mobile Edge | Clients resource Management |
| P4 | Communication-Efficient On-Device Machine Learning: Federated Distillation and Augmentation under Non-IID Private Data | Data exchange optimization |
| P5 | Efficient Training Management for Mobile Crowd-Machine Learning: A Deep Reinforcement Learning Approach | Clients resource Management |
| P6 | Two-Stream Federated Learning: Reduce the Communication Costs | Convergence acceleration |
| P7 | Adaptive Federated Learning in Resource Constrained Edge Computing Systems | Data exchange optimization |
| P8 | Federated Learning over Wireless Networks: Optimization Model Design and Analysis | Data exchange optimization |
| P9 | CMFL: Mitigating Communication Overhead for Federated Learning | Data exchange optimization |
| P10 | Distilling On-Device Intelligence at the Network Edge | Data exchange optimization |
| P11 | Federated Learning with Additional Mechanisms on-Clients to Reduce Communication Costs | Convergence acceleration |
| P12 | Towards Faster and Better Federated Learning: A Feature Fusion Approach | Convergence acceleration |
| P13 | On-Device Federated Learning via Second-Order Optimization with Over-the-Air Computation | Convergence acceleration |
| P14 | Model Pruning Enables Efficient Federated Learning on Edge Devices | Convergence acceleration |
| P15 | Robust and Communication-Efficient Federated Learning from Non-IID Data | Data exchange optimization |
| P16 | Astraea: Self-balancing Federated Learning for Improving Classification Accuracy of Mobile Deep Learning Applications | Clients resource Management |
| P17 | Communication-Efficient Federated Learning for Wireless Edge Intelligence in IoT | Data exchange optimization |
| P18 | Performance Optimization of Federated Learning over Wireless Networks | Clients resource Management |
| P19 | Communication-Efficient Federated Deep Learning With Layerwise Asynchronous Model Update and Temporally Weighted Aggregation | Data exchange optimization |
| P20 | Intermittent Pulling with Local Compensation for Communication-Efficient Federated Learning | Data exchange optimization |
| P21 | Faster On-Device Training Using New Federated Momentum Algorithm | Convergence acceleration |
| P22 | BACombo—Bandwidth-Aware Decentralized Federated Learning | Clients resource Management |
| P23 | Ternary Compression for Communication-Efficient Federated Learning | Data exchange optimization |
| P24 | Dynamic Sampling and Selective Masking for Communication-Efficient Federated Learning | Data exchange optimization |
| P25 | SAFA: a Semi-Asynchronous Protocol for Fast Federated Learning with Low Overhead | Clients resource Management |
| P26 | Federated Neuromorphic Learning of Spiking Neural Networks for Low-Power Edge Intelligence | Data exchange optimization |
| P27 | Towards Communication-Efficient Federated Learning in the Internet of Things with Edge Computing | Data exchange optimization |
| P28 | Energy-Efficient Radio Resource Allocation for Federated Edge Learning | Clients resource Management |
| P29 | Client-Edge-Cloud Hierarchical Federated Learning | Clients resource Management |
| P30 | Energy-Aware Analog Aggregation for Federated Learning with Redundant Data | Clients resource Management |
| P31 | Convergence Time Minimization of Federated Learning over Wireless Networks | Clients resource Management |

| P32 | Optimizing Federated Learning on Non-IID Data with Reinforcement Learning | Clients resource Management |
|---|---|---|
| P33 | Accelerating Federated Learning via Momentum Gradient Descent | Convergence acceleration |
| P34 | Optimal User Selection for High-Performance and Stabilized Energy-Efficient Federated Learning Platforms | Clients resource Management |
| P35 | FedPAQ: A Communication-Efficient Federated Learning Method with Periodic Averaging and Quantization | Data exchange optimization |
| P36 | Accelerating Federated Learning over Reliability-Agnostic Clients in Mobile Edge Computing Systems | Clients resource Management |
| P37 | Q-GADMM: Quantized Group Admm For Communication Efficient Decentralized Machine Learning | Data exchange optimization |
| P38 | FedMCCS: Multi Criteria Client Selection Model for Optimal IoT Federated Learning | Clients resource Management |
| P39 | FetchSGD: Communication-Efficient Federated Learning with Sketching | Data exchange optimization |
| P40 | FedAT: A Communication-Efficient Federated Learning Method with Asynchronous Tiers under Non-IID Data | Data exchange optimization |
| P41 | Lazily Aggregated Quantized Gradient Innovation for Communication-Efficient Federated Learning | Data exchange optimization |
| P42 | Toward Communication-Efficient Federated Learning in the Internet of Things With Edge Computing | Data exchange optimization |
| P43 | Group Knowledge Transfer: Federated Learning of Large CNNs at the Edge | Clients resource Management |
| P44 | CatFedAvg: Optimizing Communication-efficiency and Classification Accuracy in Federated Learning | Clients resource Management |
| P45 | Adaptive Gradient Sparsification for Efficient Federated Learning: An Online Learning Approach | Data exchange optimization |
| P46 | Communication-Efficient Federated Distillation | Data exchange optimization |
| P47 | A Trust and Energy-Aware Double Deep Reinforcement Learning Scheduling Strategy for Federated Learning on IoT Devices | Clients resource Management |
| P48 | Device Scheduling for Energy-Efficient Federated Learning over Wireless Network Based on TDMA Mode | Clients resource Management |
| P49 | Distillation-Based Semi-Supervised Federated Learning for Communication-Efficient Collaborative Training with Non-IID Private Data | Data exchange optimization |
| P50 | Time-Correlated Sparsification for Communication-Efficient Federated Learning | Data exchange optimization |
| P51 | Energy-Aware Resource Management for Federated Learning in Multi-Access Edge Computing Systems | Clients resource Management |
| P52 | FEDZIP: A Compression Framework for Communication-Efficient Federated Learning | Data exchange optimization |
| P53 | FedProf: Optimizing Federated Learning with Dynamic Data Profiling | Clients resource Management |
| P54 | Toward Resource-Efficient Federated Learning in Mobile Edge Computing | Clients resource Management |
| P55 | HeteroFL: Computation and Communication Efficient Federated Learning for Heterogeneous Clients | Clients resource Management |
| P56 | Wirelessly Powered Federated Edge Learning: Optimal Tradeoffs Between Convergence and Power Transfer | Data exchange optimization |
| P57 | Gradient Statistics Aware Power Control for Over-the-Air Federated Learning in Fading Channels | Clients resource Management |
| P58 | Accelerating Federated Learning for IoT in BigData Analytics With Pruning, Quantization andSelective Updating | Convergence acceleration |
| P59 | Communication Efficient Federated Learning with Adaptive Quantization | Data exchange optimization |
| P60 | Adaptive Batch Size for Federated Learning in Resource-Constrained Edge Computing | Convergence acceleration |
| P61 | Communication-Efficient Federated Learning with Dual-Side Low-Rank Compression | Data exchange optimization |
| P62 | Resource-Efficient Federated Learning with Hierarchical Aggregation in Edge Computing | Clients resource Management |
| P63 | Adaptive Federated Dropout: Improving Communication Efficiency and Generalization for Federated Learning | Convergence acceleration |
| P64 | To Talk or to Work: Flexible Communication Compression for Energy Efficient Federated Learning over Heterogeneous Mobile Edge Devices | Data exchange optimization |

| P65 | Adaptive Quantization of Model Updates for Communication-Efficient Federated Learning | Data exchange optimization |
| P66 | COFEL: Communication-Efficient and Optimized Federated Learning with Local Differential Privacy | Data exchange optimization |
| P67 | Energy-Efficient Federated Edge Learning with Joint Communication and Computation Design | Clients resource Management |